\documentclass[acmtog]{acmart}

\usepackage[nolist]{acronym}
\usepackage{multirow}
\usepackage{array}
\usepackage{arydshln}
\usepackage{xcolor}
\usepackage{comment}
\usepackage{pifont}% http://ctan.org/pkg/pifont
\usepackage{tabularx}
\newcommand{\cmark}{\ding{51}}%
\newcommand{\xmark}{\ding{55}}%

\begin{acronym}
    \acro  {dofs}   [DOFs]   {degrees of freedom}
    \acro  {dof}   [DOF]   {degree of freedom}
    \acro  {fov}   [FOV]   {field of view}
    \acro  {auc} [AUC]  {Area Under the Curve}
    \acro  {mpjpe} [MPJPE] {mean-per-joint-position-error}
    \acro  {mpj} [MPJPA] {mean-per-joint-position-acceleration}
    \acro  {vrar} [VR/AR] {virtual and augmented reality}
    \acro  {pck} [PCK] {Percentage of Correct Keypoints}
    \acro  {ftl} [FTL] {Feature Transform Layer}
    \acro  {dpe} [DPE] {Direct Pose Estimation Network}    
    \acro  {svd} [svd] {Singular Value Decomposition}
    \acro  {sl1} [S1] {Smooth L1 loss}
    \acro  {mse} [mse] {Mean squared error}
    \acro  {roi} [ROI] {region of interest}
\end{acronym}

\definecolor{rob_color}{rgb}{0.35,0.35,0}

\AtBeginDocument{%
  \providecommand\BibTeX{{%
    \normalfont B\kern-0.5em{\scshape i\kern-0.25em b}\kern-0.8em\TeX}}}

\copyrightyear{2022}
\acmYear{2022}
\acmYear{2022}
\setcopyright{rightsretained}
\acmConference[SA '22 Conference Papers]{SIGGRAPH Asia 2022 Conference
Papers}{December 6--9, 2022}{Daegu, Republic of Korea}
\acmBooktitle{SIGGRAPH Asia 2022 Conference Papers (SA '22 Conference
Papers), December 6--9, 2022, Daegu, Republic of
Korea}\acmDOI{10.1145/3550469.3555378}
\acmISBN{978-1-4503-9470-3/22/12}

\acmSubmissionID{170}

\newcommand{\figref}[1]{Figure~\ref{fig:#1}}

\newcommand{\secref}[1]{Section~\ref{sec:#1}}

\newcommand{\hsc}[1]{\textcolor{red}{{Shangchen: #1}}}
\newcommand{\hlinguang}[1]{\textcolor{magenta}{{linguang: #1}}}

\newcommand{\revised}[1]{{#1}}

\newcommand{\hrywang}[1]{}
\newcommand{\hcdtwigg}[1]{}
\newcommand{\hcwan}[1]{}
\newcommand{\hrc}[1]{}
\newcommand{\hbb}[1]{}
\newcommand{\hrccomment}[2]{}
\newcommand{\hrcrep}[2]{}

\begin{document}

\title{UmeTrack:  Unified multi-view end-to-end hand tracking for VR}

\author{Shangchen Han}
\email{shchhan@fb.com}
\author{Po-chen Wu}
\email{pochenwu@fb.com}
\author{Yubo Zhang}
\email{yubozhang@fb.com}
\author{Beibei Liu}
\email{beibeiliu@fb.com}
\affiliation{\institution{Meta Reality Labs} \country{USA}}

\author{Linguang Zhang}
\email{linguang@fb.com}
\author{Zheng Wang}
\email{wangz@fb.com}
\author{Weiguang Si}
\email{weiguang@fb.com}
\author{Peizhao Zhang}
\email{stzpz@fb.com}
\affiliation{\institution{Meta Reality Labs} \country{USA}}

\author{Yujun Cai}
\email{yujuncai@fb.com}
\author{Tomas Hodan}
\email{tomhodan@fb.com}
\author{Randi Cabezas}
\email{rcabezas@fb.com}
\author{Luan Tran}
\email{tranluan07@fb.com}
\affiliation{\institution{Meta Reality Labs} \country{USA}}

\author{Muzaffer Akbay}
\email{muzoakbay@fb.com}
\author{Tsz-Ho Yu}
\email{thyu@fb.com}
\author{Cem Keskin}
\email{cemkeskin@fb.com}
\author{Robert Wang}
\email{rywang@fb.com}
\affiliation{\institution{Meta Reality Labs} \country{USA}}

\renewcommand{\shortauthors}{Han et al.}

\begin{abstract}
Real-time tracking of 3D hand pose in world space is a challenging problem and plays an important role in VR interaction.
Existing work in this space are limited to either producing root-relative (versus world space) 3D pose or rely on multiple stages such as generating heatmaps and kinematic optimization to obtain 3D pose.
Moreover, the typical VR scenario, which involves multi-view tracking from wide \ac{fov} cameras is seldom addressed by these methods.  
In this paper, we present a unified end-to-end differentiable framework for multi-view, multi-frame hand tracking that directly predicts 3D hand pose in world space.
We demonstrate the benefits of end-to-end differentiabilty by extending our framework with downstream tasks such as jitter reduction and pinch prediction.
To demonstrate the efficacy of our model, we further present a new large-scale egocentric hand pose dataset that consists of both real and synthetic data. 
Experiments show that our system trained on this dataset handles various challenging interactive motions, and has been successfully applied to real-time VR applications.

\end{abstract}

\begin{CCSXML}
<ccs2012>
<concept>
<concept_id>10010147</concept_id>
<concept_desc>Computing methodologies</concept_desc>
<concept_significance>500</concept_significance>
</concept>
</ccs2012>
\end{CCSXML}

\ccsdesc[500]{Computing methodologies}

\keywords{motion capture, hand tracking, virtual reality}

%  \centering
%  \includegraphics[width=\linewidth]{figures/teaser.pdf}
%  \caption{
%  We present an end-to-end differentiable architecture for hand pose estimation designed for VR/AR headsets. The trained network can be used for direct object manipulation, UI control and driving avatars. \tom{Show a headset with highlighted cameras in the left tile and then three tiles with screenshots of different applications, emphasizing that the presented approach has been successfully deployed in these applications?}

\begin{teaserfigure}
	\begin{center}
	    \begingroup
        \setlength{\tabcolsep}{2.0pt}
        \renewcommand{\arraystretch}{0}
		\begin{tabular}{c c c c}
			\includegraphics[width=0.243\linewidth]{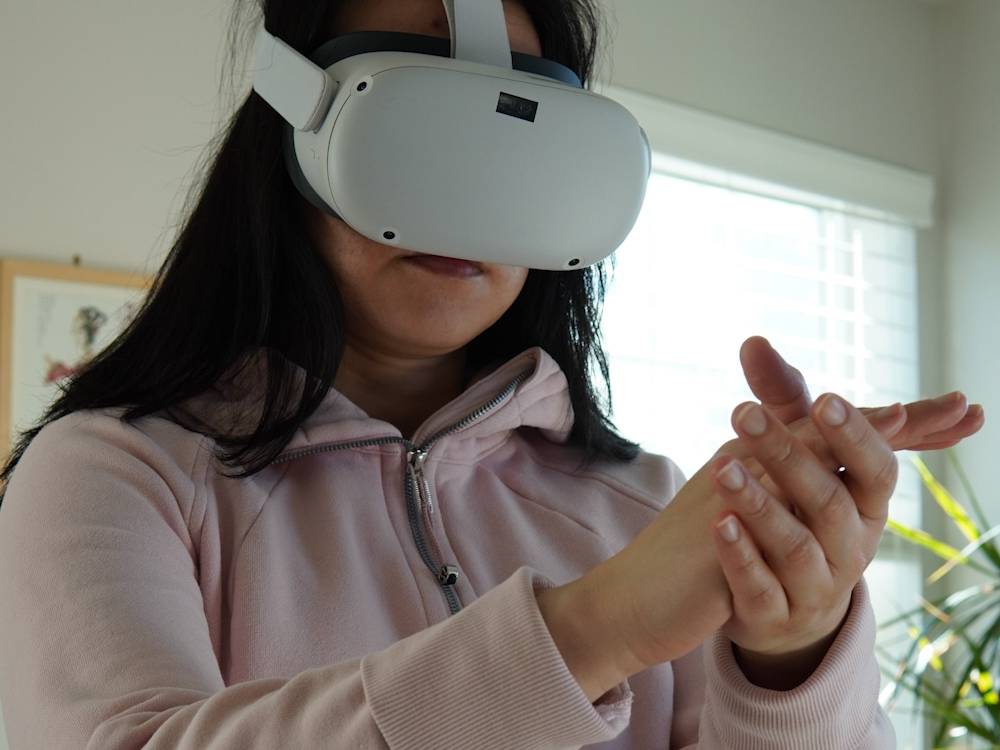} &
			\includegraphics[width=0.243\linewidth]{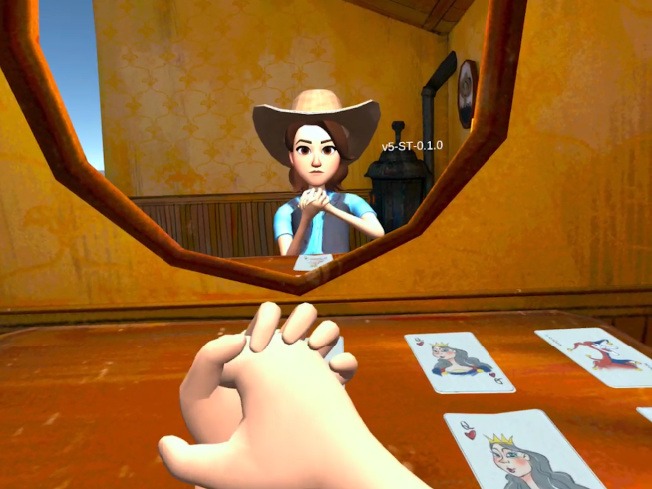} &
			\includegraphics[width=0.243\linewidth]{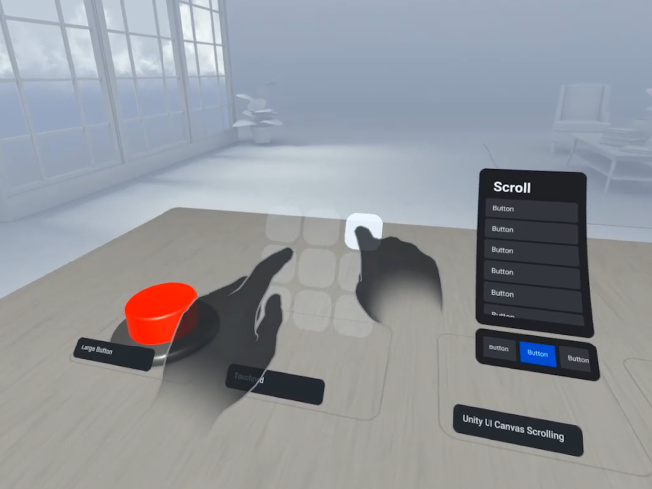} &
			\includegraphics[width=0.243\linewidth]{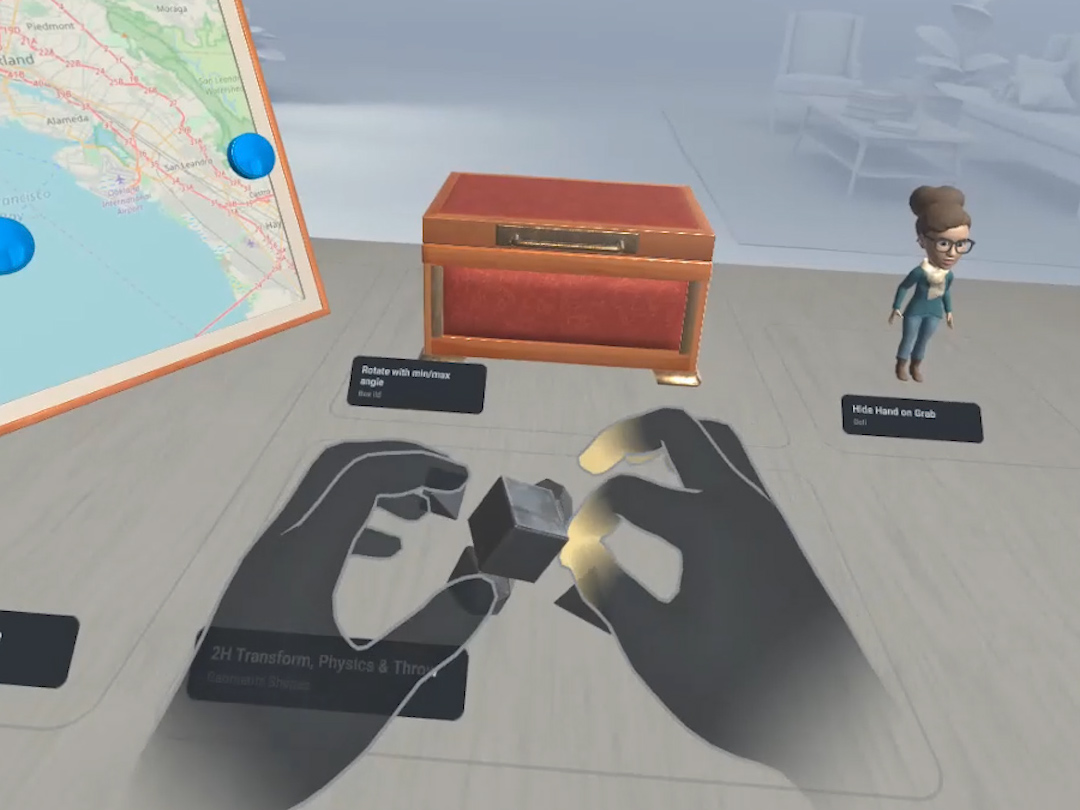} \vspace{-0.5em} \\
		\end{tabular}
		\endgroup
		\caption{\label{fig:teaser}
			The proposed hand tracking method can effectively incorporate information from multi-view image sequences captured by cameras mounted on a VR headset (left) and has been successfully deployed in various VR applications (right).
			\vspace{1.0em}
		}
	\end{center}
\end{teaserfigure}

% Figures
% 1, teaser: some interactions
% 2, pipeline: by Po-chen
% 3, visualize images generated by a pinhole crop
% 4, pck aucs on dpe_small_test and hhi_test
% 5, sample visualizations between jlp and dpe
% 6, uncertainty and error correlation plot
% 7, visualize predicted uncertainties
% 8, failure cases

\maketitle
\section{Introduction}

Commercial headsets for \ac{vrar}, including Meta Quest, Pico Neo, HTC Vive and the HP Reverb G2 deliver immersive experiences for gaming, communication, productivity and fitness. 
While the first generation of these headsets primarily relied on controllers for gaming, recent headsets have shifted towards interactions based on hand tracking to deliver a more natural experience and cater to applications outside of gaming.

Classical methods for hand tracking, e.g.~\cite{sharp2015accurate,taylor2016efficient,MeGATrack_Siggraph2020}, typically employ multiple stages, first predicting heatmaps or correspondences and then solving for the 3D hand pose with a separate optimization.
However, these multi-stage methods are supervised with proxy losses (e.g. heatmap loss) rather than the actual target metrics (e.g. 3D pose accuracy).
A recent line of research~\cite{spurr2021self,cai2018weakly, zimmermann2017learning} explores single-stage architectures that can be trained end-to-end to directly predict hand pose.  
\revised{However, these approaches only make root-relative 3D pose predictions. Many VR interactions (examples in \figref{teaser}) require absolute 3D root poses and these methods need to adopt an additional absolute root pose recovery stage to be applicable for these types of interactions.}

Moreover, the hand tracking systems deployed on VR headsets operate on multi-view image sequences captured by wide \ac{fov} RGB or monochrome cameras.  
However, an architecture that unifies multi-view fusion, temporal fusion and handling of wide \ac{fov} cameras is still missing.
\cite{MeGATrack_Siggraph2020} proposed a relative distance parameterization to handle images from wide \ac{fov} cameras but their system is designed as a multi-stage pipeline.
\cite{Zhou_dkpr, boukhayma20193d, zhou2020monocular} designed their networks for single-view hand pose estimation.
Incorporating multi-view or temporal information has been separately studied in previous work \cite{cai2019exploiting, seqhand2020, mvhm2021}, but never fully unified.

In this paper, we propose an end-to-end differentiable architecture that can predict the absolute 3D hand pose and handle both single- and multi-view temporal data from wide \ac{fov} fisheye cameras. The end-to-end differentiability allows us to optimize not only pose accuracy, but also other key aspects of the user/developer experience including jitter and pinch detection, both of which are critical to VR interaction. With an end-to-end differentiable framework, we achieve
superior jitter metric through a temporal loss and more
accurate pinch detections through a pinch loss than the stage-of-the-art multi-stage hand tracking system.

%Theoretically, an end-to-end trained network could dedicate more of its capacity to satisfy the target metrics instead of preserving intermediate information.
%However, such single-stage pipelines have been studied only for a single-view, single-frame setup, which has several limitations: (1)~depth ambiguity is inherent in single-view data, making the prediction of 3D hand pose and hand size ill-posed, (2)~multi-view fusion techniques, especially techniques for egocentric camera layouts, are under-explored, (3)~temporal consistency is not enforced in single-frame methods, leading to erratic predictions under occlusion, (4)~loss functions and evaluation metrics are designed to optimize accuracy, but higher accuracy alone does not guarantee better user experience in \ac{vrar}. In this paper, we demonstrate a novel hand pose estimation architecture that attempts to address these problems in a unified way. Our architecture effectively incorporates multi-view and multi-frame information, predicts the hand size, 3D hand pose and pose uncertainty. Moreover, this architecture is end-to-end trainable, allowing us to optimize not only for accuracy, but also other key contributors to user and developer experiences such as lower jitter and pinch gesture recognition.

As the proposed method is based on a deep neural network, a dataset for training and evaluation is essential.
To this end, we introduce a large-scale egocentric hand tracking dataset. This dataset was collected using 4 fisheye monochrome cameras featuring both real and synthetic data with large variations.
Both single-hand motions as well as challenging hand-hand interactions are included in the dataset.
The dataset also contains dedicated pinch sequences with annotated pinch events to study the interplay between action recognition and pose estimation tasks. Our proposed method trained on this dataset shows robust performance on challenging hand motions and has been successfully deployed in several VR applications (\figref{teaser}).

%\vspace{2.0ex}
\hfill \break
\noindent
This work makes the following contributions:
\begin{enumerate}
\item An end-to-end differentiable architecture that unifies multi-view fusion, temporal fusion and handling of wide \ac{fov} images while making absolute 3D hand pose predictions. This unification has only been achieved using multi-stage methods in previous work.
\item By leveraging the end-to-end differentiability, we achieve superior jitter metric through a temporal loss and more accurate pinch detections through a pinch loss than the state-of-the-art multi-stage hand tracking system.
\item A new large-scale egocentric dataset featuring single-hand motions and hand-hand interactions with 1397 real and 1397 synthetic sequences from 53 users. The dataset will be publicly released.
\end{enumerate}
\section{Related Work}

\subsection{Pose estimation using neural networks}
Many hand pose estimation methods start by predicting heat maps to
estimate 2D keypoints, typically from a single view ~\cite{iqbal2018hand,cai20203d}. To better handle self-occlusion and depth ambiguity, multi-view data is integrated through triangulation~\cite{simon2017hand} or post-inference optimization~\cite{simon2017hand, MeGATrack_Siggraph2020}. 
Recent work on multi-view fusion within the network can be achieved using latent features~\cite{epipolartransformers, Remelli_2020_CVPR, iskakov2019learnable}. 
%\paragraph{Heat map regression approaches}
% Typical multi-stage pose estimation methods start by predicting heat maps from single images to estimate 2D keypoints~\cite{iqbal2018hand, simon2017hand}.
% Post-inference optimization can be used to recover 3D keypoints or hand poses ~\cite{simon2017hand, MeGATrack_Siggraph2020}.
For instance, \citet{Remelli_2020_CVPR}~used the~\ac{ftl} to learn camera geometry-aware latent features and supervise 3D keypoint positions in an end-to-end differentiable manner. 
However, reconstructing the hand pose from keypoints usually requires an additional optimization stage.
% However, reconstructing the 3D pose from keypoints still requires an extra non-differentiable optimization. In our work, we leverage \ac{ftl} to add 3D structures in the latent features which allows us to make absolute 3D pose prediction.

%\paragraph{Direct pose regression approaches}

% Despite of the promising root-relative results generated from the end-to-end differentiable architectures~\cite{theodoridis2020cross,spurr2018cross}, directly estimating 3D pose estimation in world space in an end-to-end differentiable manner still remains challenging. %
Benefiting from the fast development of learning-based framework, many recent papers ~\cite{theodoridis2020cross,boukhayma20193d,spurr2018cross} provided promising results for the root-relative 3D hand pose from single images with end-to-end differentiable architectures. However, estimating absolute 3D pose using a single network still remains limited. 
For instance, \citet{boukhayma20193d,Kulon_2020_CVPR, seqhand2020} use a weak perspective projection camera model which inherently carries depth ambiguity. \cite{cai2019exploiting, zimmermann2017learning, mvhm2021, zhou2020monocular} predict root-relative 3D hand poses and a global alignment to ground truth is performed before making evaluations.
\citet{Moon_2019_ICCV_3DMPPE} proposed a distance-aware architecture to predict absolute root locations but it is later shown to be insufficient to give accurate depth predictions~\cite{Moon_2020_ECCV_InterHand2.6M}. Recent work \cite{cai2019exploiting, seqhand2020, mvhm2021,hasson2020leveraging} also explored using mutli-view information or temporal information to address ambiguities in single images but they are still limited to root-relative 3D pose estimation. The architecture proposed in this paper is designed to incorporate multi-view, temporal information while predicting absolute 3D hand pose.

% \cite{Spurr_BMC} showed how to apply biomechanical constraints to reduce the depth-scale ambiguity. In addition, many direct regression methods still rely on optimization-based refinement~\cite{geEtAl2019,rong2021frankmocap} or iterative refinement~\cite{pushingTheEnvelope2019,Zhang_2019_ICCV} for the best results.

\subsection{Hand pose datasets}
Table~\ref{table:dataset_stats} presents a summary of the different datasets used for hand pose estimation. \citet{Moon_2020_ECCV_InterHand2.6M} introduced a large high quality dataset with challenging hand motions labeled with both manual annotations and bootstrapping methods using multiple high resolution RGB cameras. 
 However, the dataset has minimal background and lighting variation. \citet{Freihand2019} also used a multi-view capture cage and included more background variation but the released dataset is restricted to training and evaluating single-frame, single-view pose estimation methods. 
 Due to the difficulty of obtaining high quality data with large variations, synthetic datasets \cite{zimmermann2017learning, GANeratedHands_CVPR2018} based on rendering or image-to-image translations have been proposed. 
 All the above datasets are limited to outside-in views captured by narrow \ac{fov} cameras, which are not suitable for egocentric hand tracking for \ac{vrar}.
 
 The available egocentric datasets are either instrumented with highly visible markers
 \cite{FirstPersonAction_CVPR2018} or have limited environmental variation \cite{kwon2021h}.
For collecting multi-view egocentric dataset, \citet{MeGATrack_Siggraph2020} introduced two approaches: (1) a mobile setup equipped with a single depth sensor where ground truth is obtained using a depth based hand tracker, and (2) a lab setup with many motion capture cameras where ground truth is obtained with a marker-based hand tracker~\cite{MocapHT_Siggraph2018}. Annotations of both approaches are obtained automatically, making them suitable for large scale data collections. However, the ground truth quality of (1) is limited by the single-view depth based hand tracker (which struggles with the ambiguity of hand-hand occlusions) and (2) contains limited environment and lighting variations. In this work, we adopt the same marker-based hand tracker for uncompromised ground truth but use synthetic data to improve environment and lighting variations. Altogether, we contribute the largest and most diverse egocentric multi-view, multi-frame dataset to date (as shown in Table~\ref{table:dataset_stats}).
 
\begin{table}[]
\centering
\caption{Comparing the proposed datasets with existing datasets.}
%\small
\vspace{-3mm}
\resizebox{\linewidth}{!}{%
\setlength{\tabcolsep}{3pt}
\begin{tabular}{@{}lrcccccc@{}}
% \begin{tabularx}{\hsize}{Xrcccccc}
\toprule
\textbf{Dataset} & \textbf{\# frames} & \textbf{ego} & \textbf{bg variation} & \textbf{markerless} & \textbf{real} & \textbf{large fov} & \textbf{hand-hand} \\
\midrule 
FreiHAND \shortcite{Freihand2019} & 37K & \xmark & \cmark & \cmark & \cmark & \xmark & \xmark \\ 
HOnnotate \shortcite{Hampali_2020_CVPR} & 78K & \xmark & \xmark & \cmark & \cmark & \xmark & \xmark \\
InterHand2.6M \shortcite{Moon_2020_ECCV_InterHand2.6M} & 2,590K & \xmark & \xmark & \cmark & \cmark & \xmark & \cmark \\ 
GANerated Hands \shortcite{GANeratedHands_CVPR2018} & 331K & \xmark & \cmark & \cmark & \xmark & \xmark & \xmark \\
H2O \shortcite{kwon2021h} & 571K & \cmark & \xmark & \cmark & \cmark & \xmark & \xmark \\
FPHA \shortcite{FirstPersonAction_CVPR2018} & 105K & \cmark & \cmark & \xmark & \cmark & \xmark & \xmark \\
\midrule 
Ours (real) & 839k & \cmark & \xmark & \xmark & \cmark & \cmark & \cmark \\ 
Ours (synth) & 839k & \cmark & \cmark & \cmark & \xmark & \cmark & \cmark \\
\bottomrule 
\end{tabular}
}
\label{table:dataset_stats}
\end{table}
%  https://dex-ycb.github.io/assets/chao_cvpr2021.pdf
% https://arxiv.org/pdf/2104.11181.pdf
\begin{figure*}[t]
 \centering
 \includegraphics[width=1\linewidth]{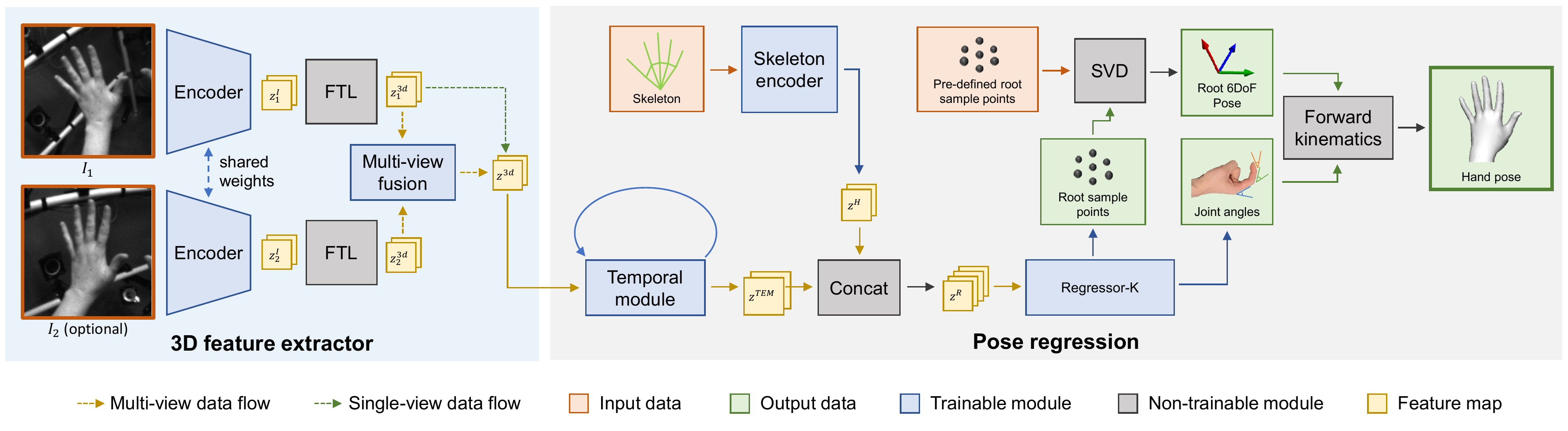}
  \caption{This figure shows our architecture for the case of a known hand skeleton. The \textbf{3D feature extractor} block can take either single-view or multi-view input data and produce 3D features $z^{3d}$ via a feature transform layer (FTL).
The \textbf{pose regression} block first generates $z^{R}$ which contains 3D information, temporal context and skeleton features. The known-skeleton regressor (Regressor-K) takes $z^{R}$ as input and predicts the absolute 3D hand pose. The whole network can be trained end-to-end as all the modules and operators are differentiable.}
  \label{fig:system_overview}
\end{figure*}

\section{Method}
\paragraph{Preliminaries}
Our hand skeleton $H$ consists of 20 joints that can be articulated by a 20 dimensional joint angle vector $\theta$. Each joint is defined by a joint position and rotation axis. The global transformation of the hand is represented by the root transformation $T_H$ consisting of 6 \ac{dofs}. With hand skeleton and predicted joint angles and root transformation, we can animate the corresponding hand mesh via linear blend skinning (LBS).
%Each skeleton is paired with a hand mesh which can be used for rendering purpose.

We adopt the multi-camera layout using wide \ac{fov} fisheye cameras placed on a VR headset by \citet{MeGATrack_Siggraph2020}. In this layout, a hand can be seen by a single camera or multiple cameras depending on the hand location.
As a result, we designed our network to jointly handle single-view and multi-vew input data.
For each frame, the input to our model is either single-view or multi-view images where the \ac{roi} around a hand is provided.

For real-world usage, our model supports both known and unknown hand skeletons. If a user has a hand skeleton generated in advance (i.e. by a scanning system), our model is capable of utilizing the skeleton information and the output is $\{\theta, T_H\}$ in this case.
The corresponding architecture is shown in \figref{system_overview}.  % In VR applications in real-world, we commonly have no prior knowledge of the hand skeletons for new users.
% To support this case, 
For new users without prior knowledge of hand skeletons, we calibrate the unknown hand skeletons by predicting hand scale from multi-view images and the model output becomes $\{\theta, T_H, H\}$.

In the following sections, \secref{cropping} describes how we prepare input images for our model.
\secref{known_hand_model} gives an overview of our model when a known hand skeleton is provided. 
\secref{unkown_hand_model} details how skeleton calibration is performed when the user's hand skeleton is unknown.
\secref{losses} introduces the loss terms we use for training our model.

\subsection{Perspective cropping for input images}
\label{sec:cropping}
\begin{figure}
 \centering
 \includegraphics[width=\linewidth]{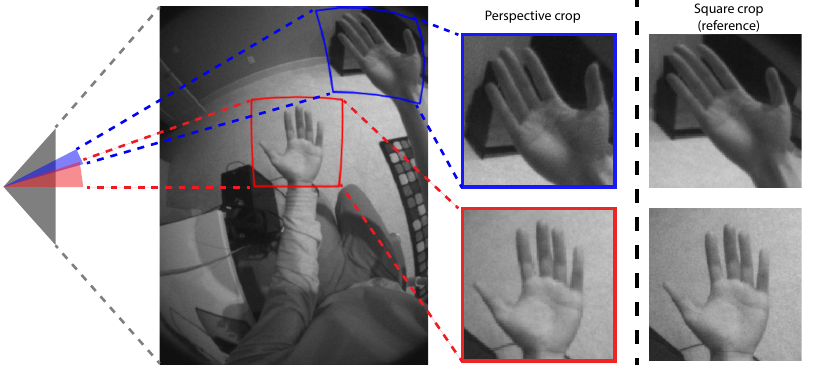}
  \caption{The perspective cropping method generates a virtual perspective camera for each hand. In this figure, each camera is represented with a different color: grey for original camera, red for left hand virtual camera and blue for right hand virtual camera. The solid lines around each hand outline the crop boundaries. On the right, we show the square crops commonly used in previous methods for comparison. Note that the square crop of the right hand is visually more distorted whereas perspective cropping can correct this distortion.}
  \label{fig:resample}
\end{figure}

Most existing works \cite{MeGATrack_Siggraph2020} utilize a square crop around a hand region as the input. Unlike images captured by a physical camera, these cropped images essentially lose the corresponding camera geometry information (i.e. intrinsics and extrinsics), making them infeasible for directly estimating absolute 3D pose.  
To tackle this issue, inspired by the effectiveness of perspective cropping~\cite{Remelli_2020_CVPR, Yu2021PCLsGN},
we adopt the same cropping approach (\figref{resample}) for generating input images for our model.
Specifically, given a \ac{roi} in the original image, we first create a virtual perspective camera at the same location as the original camera. The extrinsics matrix is constructed such that the virtual camera's z-axis points at the center of the \ac{roi}.
% If the \ac{roi} contains a right hand, we mirror along the x-axis to produce a flipped image, mimicking a left hand.
At train time, the virtual camera’s intrinsics matrix is set such that all the ground truth hand keypoints are projected within the image boundary. At inference time when ground truth is not available, we rely on the previously tracked pose to estimate the intrinsics matrix. Once the virtual camera is created, the warping technique described by \cite{Yu2021PCLsGN} can be used to generate the cropped image. By using perspective cropping, we achieved 2 goals: (1) distortions from fisheye images can be corrected (as shown in \figref{resample}); (2) the virtual camera provides essential geometry information for 3D space, which enables direct hand scale prediction and absolute 3D pose prediction.

% Common approaches use a square crop around a hand region as the input for a network.
% However, unlike images captured by a physical camera, cropped images don't have any coppresonding camera geometry (i.e. intrinsics and extrinsics). As a result, there is no well defined 3D space to begin with which makes absolute 3D pose estimation infeasible.
% In our work, we adopt a perspective cropping method (\figref{resample}), which has been used for human pose estimation \cite{Remelli_2020_CVPR, Yu2021PCLsGN}. Given a \ac{roi} in the original image, we first create a virtual camera at the same location as the original camera. The extrinsics matrix is constructed such that the virtual camera's z-axis points at the center of the \ac{roi}.
% If the \ac{roi} contains a right hand, we apply a mirror transform to the extrinsics matrix so that a left hand image can be generated after image warping.
% The virtual camera's focal length is set such that the entire hand region is projected within the image boundary.
% Once the virtual camera is created, the warping technique described by \citet{Yu2021PCLsGN} can be used to generate the cropped image.
% By creating a virtual camera, we achieved 2 goals: (1) distortions from fisheye images are corrected (as shown in \figref{resample}); (2) the virtual camera geometry provides essential information for direct hand size prediction from multi-view data as well as absolute 3D pose prediction.

\subsection{Architecture with known hand skeleton}
\label{sec:known_hand_model}
In this section, we discuss the network architecture with known hand skeleton, as is shown in Figure~\ref{fig:system_overview}.
%For new users without pre-defined hand skeletons, we designs a slightly different regresssor module for skeleton calibration and the skeleton encoder is not used for this architecture, which will be elaborated in Section~\ref{sec:unkown_hand_model}. 
% Note that the main difference between known and unknown hand skeleton lies in the Regressor module, which will be demonstrated in Section~\ref{sec:unkown_hand_model}

% \label{sec:network}
%\figref{system_overview} gives an overview of the proposed method given a known hand model. 
\paragraph{3D feature extractor}
Given the input cropped images
$\{I_{i}\}_{i=1}^n$, 
%$\{I_{i,t}|i=1...N, t=1...T\}_{i=1}^n$, 
where $N$ represents the number of hand crops, we first feed them into a fully-convolutional encoder to generate the latent features $z_{i}^I$.  %$\{z_{i}^I\}_{i=1}^n$.
% We denote the input images as $\{I_i\}_{i=1}^n$ where $n$ represents the number of hand crops. Each cropped image is first passed into a fully-convolutional encoder to yield latent features $\{z_i^I\}_{i=1}^n$. 
Here $z_{i}^I$ is extracted from the image space without knowing the camera geometry, which is insufficient for scale prediction or absolute 3D pose prediction. 
To incorporate 3D structures into the network, we leverage \ac{ftl}~\cite{Remelli_2020_CVPR, Worrall17} to generate camera-geometry-aware latent features:
% More concretely, \ac{ftl} first reshapes the feature maps into a 3D point set, and then applies a target transform to each point and reshape the transformed points back to feature maps. In our work, we apply \ac{ftl} as follows:
\begin{align}
z_{i} ^ {3d} = \text{FTL} (z_{i} ^I|T_{i, 1} * K_i ^{-1}).
\end{align}
Here, the image features $z_{i} ^I$ is first unprojected to 3D space using the virtual camera intrinsics matrix $K_i$, and then transformed from the 3D space of virtual camera $i$ into the 3D space of the reference camera. In practice, we set virtual camera 1 as the reference camera and $T_{i, 1}$ denotes the transform from camera $i$ to camera $1$ (so $T_{1,1}$ is identity in particular).
By doing so, we integrate the 3D structural information into the image features and the transformed features $\{z_{i} ^ {3d}\}_{i=1}^n$ are all in the same 3D space.

Our model can handle both single-view ($n=1$) and multi-view ($n>1$) data.
We denote the output of \textit{3D feature extractor} as $z^{3d}$.
For single-view input, the features can be directly set with: $z ^ {3d} = z_1^{3d}$. For multi-view input, a multi-view fusion module $MVF$ is deployed to fuse features from different camera views:
\begin{align}
z ^ {3d} = MVF(concatenate(z_i^{3d}, ..., z_n^{3d})), i \in [1,n].
\end{align}
The output $z ^ {3d}$ of MVF has the same shape as $z_i^{3d}$ so that the following modules in our model that consume $z^{3d}$ as input can be agnostic to the number of input images.
Since $z_i^{3d}$ carries the inter-camera relationships thanks to \ac{ftl}, $MVF$ is capable of performing learnable triangulations in the feature space ~\cite{iskakov2019learnable} and is the key to our ability to predict hand scale (See Section~\ref{sec:unkown_hand_model} for more details).

% $MVF$ is designed to output $z ^ {3d}$ that has the same shape as $z_i^{3d}$.
% Since $z_i^{3d}$ carries the inter-camera relationships thanks to \ac{ftl}, $z ^ {3d}$ produced by $MVF$ is able to resolve depth ambiguity.
\paragraph{Temporal Module}
%\subparagraph{Temporal Module}
Given a set of (single-view or multi-view) spatial features $z ^ {3d}$, we start the pose regression step with a temporal module $TEM$, which fuses $z ^ {3d}$ with the temporal context: $z^{TEM}=TEM(z^{3d}, h_{TEM})$. The temporal module is a recurrent neural network with hidden states denoted as $h_{TEM}$. It aims to learn temporal consistency, and serves as the key to handling hands under severe occlusions.

\paragraph{Skeleton Encoder}
The task of pose regression is ambiguous without the skeleton data: (1) if the input is a single image, depth ambiguity cannot be
resolved without knowing the hand scale, and (2) joint angles need to be coupled with joint positions and rotation axes to define hand
articulations. When having the prior knowledge of the skeleton data, we introduce a skeleton encoder $SE$ to implicitly incorporate hand skeleton information into the network. Specifically, we turn the joint positions (3-dimentional) and rotation axes (3-dimensional) at the rest hand pose into a feature vector.
With 20 joints, we can obtain a 120 dimentional feature vector.
The skeleton encoder SE takes a skeleton $H$ as input, performs featurization and encodes the skeleton features into feature maps: $z ^ {H} = SE(H)$.

\paragraph{Regressor-K}
The last learnable module is Regressor-K ("K" represents "\textbf{k}nown hand skeleton"). The input to Regressor-K is:
\begin{align}
z ^ {R} = concatenate (z ^ {TEM}, z ^ {H}),
\end{align}
where the concatenated feature $z ^ {R}$ contains 3D space information, temporal context and hand skeleton data. Regressor-K takes $z^{R}$ as input and predicts 20 joint angles $\hat{\theta}$ and 3D root points $\hat{v}$ which encodes the root transform $\hat{T}_{H, 1}$ in the reference camera space.
To decode $\hat{T}_{H, 1}$ from $\hat{v}$, we use Singular Value Decomposition~\cite{SorkineRabinovich:SVD-rotations:2016} to align pre-defined 3D root points in the local hand coordinate system to the predicted 3D root points $\hat{v}$.
More details on decoding $\hat{T}_{H, 1}$ can be found in the supplementary material.
Then we recover the root transform $\hat{T}_H$ into the world coordinate system using the extrinsics of the reference camera. The output $\{\hat{\theta}, \hat{T}_H\}$ can be used to render the hand and drive applications in VR.

\subsection{Calibration for unknown hand skeleton}
\label{sec:unkown_hand_model}
\begin{figure}
 \centering
 \includegraphics[width=\linewidth]{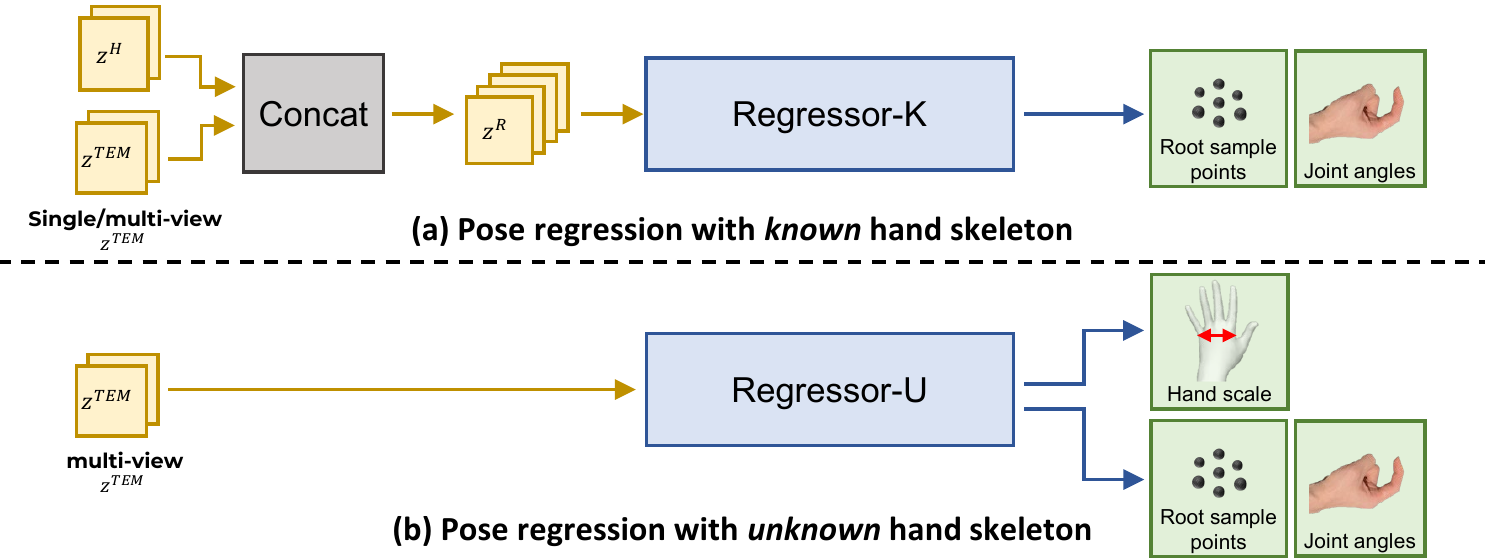}
  \caption{we provide two regressors for (a) known and (b) unknown hand skeletons. 
  %The proposed architecture contains two regressor modules. 
  (a) Regressor-K is used when a known hand skeleton is provided. Regressor-K is capable of handling $z^{TEM}$ computed from either single-view or multi-view data and requires a known hand skeleton for computing $z^{H}$. (b) Regressor-U is used when the hand skeleton is unknown. Regressor-U requires $z^{TEM}$ being computed from multi-view data and predicts an additional hand scale that can be used to generate the estimated skeleton.}
  \label{fig:regressor}
\end{figure}

We pose the hand skeleton calibration problem as a hand scale calibration task similar to \cite{MeGATrack_Siggraph2020}.
The calibrated hand scale can be used to scale a reference hand skeleton. 
Calibration of more parameters for a hand similar to \cite{MANO} will be a future extension of the proposed method.
For hand scale calibration, our method relies on Regressor-U ("U" represents "\textbf{u}nknown hand skeleton").
The differences between Regressor-K and Regressor-U are shown in \figref{regressor}.
As shown in  \figref{regressor} (b), Regressor-U requires the temporal features $z ^ {TEM}$ being computed from multi-view data since single-view data inherently carries scale ambiguity, making scale calibration an ill-posed problem.
Regressor-U outputs an additional hand scale estimation which can be used to generate the calibrated hand skeleton $\hat{H}$. The final output of Regressor-U is $\{\hat{\theta}, \hat{T}_H, \hat{H}\}$.
During training, both Regressor-K and Regressor-U can be trained jointly in an end-to-end manner.
During inference, our model can dynamically decide which regressor to use depending on whether the skeleton data is provided.

\subsection{Loss functions}
\label{sec:losses}
The network is trained using a combination of three loss terms: pose loss, temporal loss and pinch loss. 
\paragraph{Pose loss} The pose loss consists of a set of regression losses to supervise the hand poses:
\begin{align}
L_{pose} = & \sum_{j}^J {||p_{j} (\theta, T_H, H) - p_{j} (\hat{\theta}, \hat{T}_H, \hat{H})||_1}  + \\
& \lambda_{\theta} ||\theta - \hat{\theta}||_1 + \lambda_w ||w(T_H) - w(\hat{T}_H)||_1,
\end{align}
where $p_{j}(.)$ is a function that computes the $j$-th keypoint position using forward kinematics and $w$ is a function that gets the translation component from $T_H$. We use the hat symbol to denote the network prediction. In the case a hand skeleton is known, $\hat{H}$ is the same as the groud truth skeleton $H$. If the hand skeleton is unknown, $\hat{H}$ is predicted by Regressor-U. We set the loss weights $\lambda_{\theta} = 0.05, \lambda_w = 0.5$ in our experiments.

\paragraph{Temporal loss} The temporal loss enforces smoothness of the predicted pose sequence by penalizing acceleration:
\begin{align}
L_{temp} = \sum_{t}^T (|| acc(\hat{\theta}, t))||_1 + ||acc(\hat{T}_H, t)||_1) 
\end{align}
where $acc(x, t) = x_{t+1} + x_{t-1} - 2 x_t$ is a function computing the acceleration of a given signal $x$ at frame $t$. Similar losses are harder to incorporate for previous end-to-end methods that predict root-relative hand poses since root transformation is unavailable in those methods.

\paragraph{Pinch loss} Reliable detection of pinch is critical because pinch is used for selection in VR interaction.  
The design of our pinch loss is based on the observation that the distance between the thumb and index fingertips is small while pinching:
\begin{align}
L_{pinch} = l \cdot \text{min} (d(\hat{\theta}, \hat{T}_H, \hat{H})-  \epsilon_1,0) + \\
(1- l) \cdot \text{min} (\epsilon_2 - d(\hat{\theta}, \hat{T}_H, \hat{H}), 0)),
\end{align}
where $l$ is an annotated binary pinch label and $d$ is a function that computes distance between the thumb and index fingertips. $\epsilon_1$ is a distance threshold that corresponds to a high chance of a pinch event. $\epsilon_2$ is a ``safe" distance threshold that means a pinch event is unlikely to happen. $\epsilon_1$ and $\epsilon_2$ are chosen to be 0.01 and 0.02 meters respectively.

The final loss is a linear combination of the above losses:
\begin{align}
L = L_{pose} + \lambda_{t} L_{temp} + \lambda_{p} L_{pinch},
\end{align}
where we set weights $\lambda_{t} = 0.05, \lambda_p=0.4$ in experiments.
\section{Dataset}
For training our pipeline in a supervised manner, a large scale multi-view egocentric dataset with challenging interactive hand motion is required. As seen in Table~\ref{table:dataset_stats}, existing datasets for hand tracking tasks are either infeasible for egocentric viewpoints or lack the challenging hand-hand motions, which are crutial for VR applications. To this end, we propose a new real-world large-scale egocentric dataset, with 1397 sequences for 53 users. 
%We adopt the same marker-based hand tracker as~\cite{MeGATrack_Siggraph2020} for uncompromised ground truth and render synthetic data to improve environment and lighting variations. 
For each user, we get the ground truth hand skeleton and mesh from a scanning system. Each sequence has 15 seconds and is captured at 30fps. Each frame contains 4 VGA images captured from 4 wide \ac{fov} monochrome cameras on a VR headset.
We set up a motion capture system with 36 cameras to track marker locations. Markers (3mm) were placed on each user's hands and a marker based hand tracker \cite{MocapHT_Siggraph2018} was used to obtain the ground truth motions. %Certain sequences may contain sporadic tracking failures and we leave those frames as unlabeled. 
In each frame, a hand with ground truth label could appear in 1 or 2 views.

The dataset is divided into two protocols: a \textit{separate-hand} protocol focusing on individual hand motions and a \textit{hand-hand} protocol focusing on inter-hand interactions. Within the \textit{separate-hand} protocol, 192 sequences contain pinch motion. Pinch events are manually annotated, resulting in 38003 pinch labels.

In addition to the real data, we rendered a synthetic dataset with the same hand motions as the real dataset using the Unity game engine. 
Each sequence is rendered using a different background and each frame is rendered using a different lighting configuration. As a result, the synthetic dataset provides much larger variations in environment and lighting. Samples of the real and synthetic data can be found in our supplementary video. As shown in Table~\ref{table:dataset_stats}, this dataset is the largest egocentric dataset to our knowledge dedicated to hand tracking in VR.
\section{Evaluation}
\label{sec:evaluation}

To evaluate the accuracy of our system, we use \ac{mpjpe} which computes the average 3D Euclidean distance in millimeters between the estimated and ground truth keypoints in world space. We use the \ac{mpj} metric to measure tracking jitter similar to ~\cite{MeGATrack_Siggraph2020} using the following equation:
\begin{align}
mpjpa (\hat{\theta}, \hat{T}_{H}, \hat{H}) = \frac{1}{T \cdot J} \sum_t^T \sum_j^J ||acc( p_j (\hat{\theta}, \hat{T}_{H}, \hat{H}), t)||_2
\end{align}
Lower \ac{mpjpe} indicates better accuracy and lower \ac{mpj} indicates less jittery tracking. 
We compare our method to the state-of-the-art multi-stage hand tracking method for VR by \cite{MeGATrack_Siggraph2020} to understand the benefit of end-to-end differentiability brought by our pipeline.

\subsection{Implementation details}
\paragraph{Data augmentation} We perform data augmentation by perturbing the camera intrinsics and extrinsics during training: (1) adding noise to the look-at direction when creating extrinsics of the virtual camera, (2) randomly applying in-plane rotation to the camera extrinsics, and (3) random scaling to the focal lengths. Note that we can't use the commonly used affine transforms to augment input images since they would cause a mismatch between virtual camera parameters and image data.

We implemented our method in PyTorch \cite{paszke2017automatic}. For fair comparisons with ~\citet{MeGATrack_Siggraph2020}, we use the same input resolution ($96\times96$ monochrome image) and the same backbone for image encoder. All the modules in our model are jointly trained in an end-to-end manner using 9 GPUs with a batch size of 144. We first train the network for 200 epochs using the Adam optimizer with a learning rate of 0.0002 without temporal or pinch loss. The network is then trained for another 200 epochs with all the loss terms enabled. 
Inference with our model using 4 images (both hands are seen by 2 cameras) takes $\sim$10ms on a PC with a NVIDIA GTX 2080 Super.
We provide more details on the layers used for each module in the supplementary material.

Empirically, we found that a network trained on synthetic data shows better robustness to challenging environments whereas a network trained on real data gives better metrics on the test split of the real dataset. To leverage the benefits of both, we train both our method and the keypoint network by \cite{MeGATrack_Siggraph2020} on the combined real and synthetic data for fair comparison. Metrics are reported on the test split of the real data.

\subsection{Ablation study}
In this section, we perform ablation study on the modules and loss functions. Pinch loss is not included and will be discussed later in section~\ref{sec:pinch_eva}. 
% later when we evaluate pinch detection accuracy.
We compare results under \ac{mpjpe} and \ac{mpj} metrics on the \textit{separate-hand} protocol, which are summarized in Table~\ref{table:ablation}.
Without \ac{ftl}, our model can barely learn to predict 3D hand pose and the corresponding model performs the worst (49.8mm \ac{mpjpe}).
The skeleton encoder extracts critical hand scale and joint features, without which, the network accuracy degrades (13.4mm \ac{mpjpe}).
Without the temporal module, the network loses the temporal context and performs slightly worse in \ac{mpjpe} and much worse in \ac{mpj}.
Adding $L_{temp}$ without the temporal module improves \ac{mpj} (5.10 to 4.39) but leads to noticeable degradation in \ac{mpjpe} (9.5 mm to 10.0mm).
When using the full model, the model trained with $L_{temp}$ shows slightly degraded \ac{mpjpe} (9.4mm vs. 9.3mm) but performs much better in \ac{mpj} metric (2.61 vs. 3.52) than the model trained without $L_{temp}$.
\revised{To further validate the benefit of $L_{temp}$, we compare the model trained with $L_{temp}$ to using a one-euro filter \cite{Casiez20121F} for post-processing. The one-euro filter parameters were tuned to produce identical \ac{mpj} metric and it performed much worse in \ac{mpjpe} metric (10.1mm vs. 9.4mm).}
Based on these observations, we consider $L_{temp}$ brings enough benefit to \ac{mpj} metric and we adopt the full model trained with both $L_{pose}$ and $L_{temp}$ for hand tracking task.

\begin{figure}
 \centering
 \includegraphics[width=\linewidth]{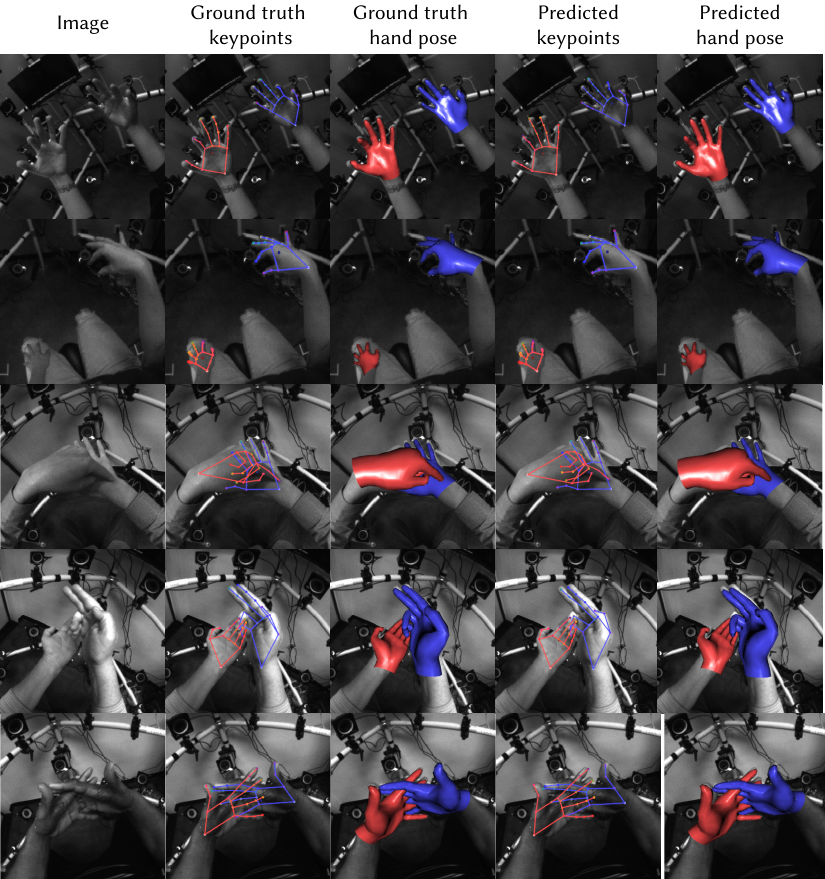}
  \caption{Qualitative results on our evaluation dataset.}
  \label{fig:qualitative}
\end{figure}

\begin{figure}
 \centering
 \includegraphics[width=0.95\linewidth]{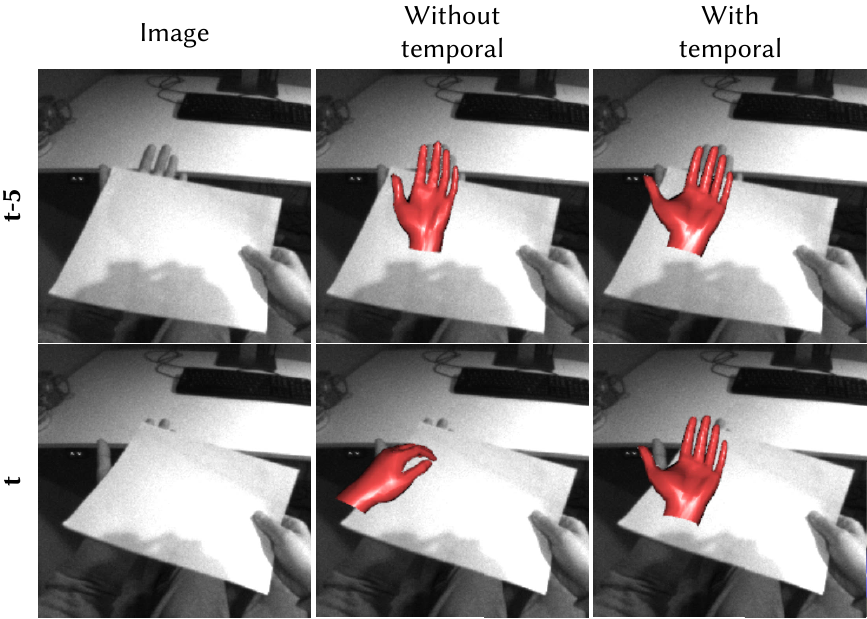}
  \caption{The model without temporal module predicts a reasonable pose at frame (t-5) but completely fails at frame t due to severe occlusion. In contrast, the model with temporal module predicts a plausible hand pose at frame t by leveraging the temporal context. 
  }
  \label{fig:compare_temporal}
\end{figure}

\figref{qualitative} visualizes some sample results using our method. In particular, hands that are severely occluded by each other can be reasonably tracked (row 3, 5 in \figref{qualitative}).
Another observation we made was the temporal module is the key to handling challenging occlusions by another object.
To show this, we provide an example in \figref{compare_temporal} where the full model is able to maintain a plausible pose when the hand is occluded by a paper whereas the model without the temporal module completely fails.
\begin{table}[t]
\caption{Ablation study for different modules and loss functions. For each model, "\textbf{|}" separates the model architecture (left) and loss functions used for training (right). Model annotated with "(One-Euro)" refers to using one-euro filter to post-process the tracked poses. Best model is highlighted in bold.}
\begin{center}
\label{table:ablation}
%
%\begin{tabular}{l|ccccc}
%
%\hline
%model & \ac{mpjpe}(gt) & \ac{mpj}(gt) & \ac{mpjpe}(s) & \ac{mpj}(s) \\
%\hline
%no \ac{ftl} & 49.8 & 2.31 & 59.4 & 2.21 \\
%no skeleton & 13.4 & 1.53 & 13.4 & 1.53 \\
%no temporal & 9.5 & 1.83 & 12.6 & 1.84 \\
%no temporal, jitter loss & 10.0 & 1.65 & 12.5 & 1.64 \\
%no jitter loss & 9.3 & 1.36 & 11.1 & 1.35 \\
%full model & 9.3 & 1.17 & 11.5 & 1.15 \\
\renewcommand{\arraystretch}{0.9}
\begin{tabular}{l|cc}
\hline
model & \ac{mpjpe} & \ac{mpj}\\
\hline
% dpe_no_ftl: f314625871 (f343246536)
w/o \ac{ftl} \textbf{|} $L_{pose}$ & 49.8 & 6.29 \\
% dpe_no_f_skel: f316581878 (f343246664)
w/o skeleton encoder \textbf{|} $L_{pose}$ & 13.4 & 4.07 \\
% dpe_no_temporal_no_jit: f314937406 (f343246592)
w/o temporal module \textbf{|} $L_{pose}$ & 9.5 & 5.10 \\
% dpe_no_temporal_w_acc: f343243601 (f343243601)
w/o temporal module \textbf{|} $L_{pose}$ + $L_{temp}$ & 10.2 & 4.39 \\
% dpe_no_jit: f313197936 (f343079657)
full model \textbf{|} $L_{pose}$ & \textbf{9.3} & 3.52 \\
% dpe_no_jit 1euro5/5/1: f313197936 (f366663758)
\revised{full model \textbf{|} $L_{pose}$ (One-Euro) }& \revised{10.1} & \revised{2.67} \\
% dpe_w_acc (image_noise): f341717862 (f341717862)
\textbf{full model |} $\mathbf{L_{pose} + L_{temp}}$ & 9.4 & \textbf{2.61} \\

% Older models
% dpe_no_temporal_w_jit (old): f315230762
% dpe_w_jit (old): f312786193
\hline
\end{tabular}
\end{center}
\end{table}

\subsection{Comparison to previous multi-stage method}
\label{sec:evaluation:compare_to_previous}
\begin{table*}[]
\caption{Comparison with the multi-stage method~\cite{MeGATrack_Siggraph2020} on \textit{separate-hand} and \textit{hand-hand} protocols. For each of our models, loss functions used for training are specified after "\textbf{|}".}
\begin{center}
\label{table:compare_previous}
\renewcommand{\arraystretch}{0.9}
\begin{tabular}{lccccccccc}
\hline
\multicolumn{1}{c}{\multirow{ 3}{*}{Method}} & \multicolumn{4}{c}{Known hand skeleton} & \multicolumn{4}{c}{Unknown hand skeleton} \\
\cmidrule(lr){2-5} \cmidrule(lr){6-9}
  & \multicolumn{2}{c}{\textit{separate-hand}} & \multicolumn{2}{c}{\textit{hand-hand}} & \multicolumn{2}{c}{\textit{separate-hand}} & \multicolumn{2}{c}{\textit{hand-hand}} \\
\cmidrule(lr){2-3} \cmidrule(lr){4-5} \cmidrule(lr){6-7} \cmidrule(lr){8-9}
 & \ac{mpjpe}  & \ac{mpj} & \ac{mpjpe} & \ac{mpj}& \ac{mpjpe}  & \ac{mpj} & \ac{mpjpe} & \ac{mpj}\\
\hline
% mf-jlp no_contact: f313179363 (f343079539)
\cite{MeGATrack_Siggraph2020} & 9.9 & 3.48 & 10.8 & 3.33 & 12.9 & 3.46 & 13.6 & 3.33 \\
\hdashline
% dpe_no_jit: f313197936 (f343079657)
% Ours \textbf{|} $L_{pose}$              & \textbf{9.3} & 3.52 & 10.6 & 3.77  & \textbf{11.1} & 3.49 & 12.4 & 3.74 \\
% dpe_w_acc (image_noise): f341717862 (f341717862)
Ours \textbf{|} $L_{pose}$ + $L_{temp}$     & \textbf{9.4} & \textbf{2.61} & \textbf{10.5} & 2.73 & \textbf{11.2} & \textbf{2.57}  & \textbf{12.0} & 2.69 \\
% dpe_w_acc l_pinch0.4 (image_noise): f344158181 (f344158181)
Ours \textbf{|} $L_{pose}$ + $L_{temp}$ + $ L_{pinch}$                        & \textbf{9.4} & 2.65 & 10.6 & \textbf{2.68} & 11.4 & 2.60  & 12.2 & \textbf{2.65} \\
\hline
\end{tabular}
\end{center}
\end{table*}

\begin{figure}
 \centering
 \includegraphics[width=\linewidth]{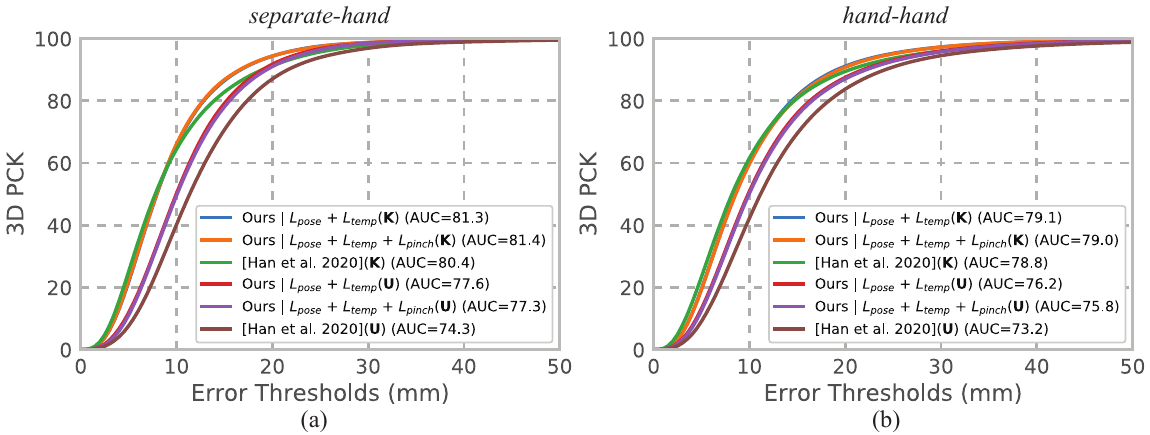}
  \caption{PCK curves on \textit{separate-hand} and \textit{hand-hand} protocols. Legends including ``(\textbf{K})'' refer to evaluation for \textbf{k}nown hand skeletons and legends with ``(\textbf{U})'' refer to evaluation for \textbf{u}nknown hand skeletons.}
  \label{fig:pck_curves}
\end{figure}
\paragraph{Known hand skeleton} We compare our model trained with both $L_{pose}$ and $L_{temp}$ to ~\citet{MeGATrack_Siggraph2020} using the ground truth skeletons provided by the dataset.
Metrics on \textit{separate-hand} and \textit{hand-hand} protocols are reported in Table~\ref{table:compare_previous}.
On both protocols, our method outperforms ~\cite{MeGATrack_Siggraph2020} in \ac{mpjpe} (9.4mm vs. 9.9mm on \textit{separate-hand}; 10.5mm vs 10.8mm on \textit{hand-hand}). 
More details on the performance of each method at different error thresholds can be found in \figref{pck_curves}(a) and (b).
In particular, in \figref{pck_curves}(a), the curve for our method crossed with the curve for ~\cite{MeGATrack_Siggraph2020} at 9.5mm error threshold. This suggests our method produces fewer outliers (error > 9.5mm) at the cost of degraded capability in precise localization (error < 9.5mm). A similar phenomenon can also be seen in \figref{pck_curves}(b).
With respect to the \ac{mpj} metric, our method consistently outperforms ~\citet{MeGATrack_Siggraph2020} (2.61 vs. 3.48 on \textit{separate-hand}; 2.68 vs. 3.33 on \textit{hand-hand}).
%particularly around fingertips as shown in \figref{pck_curves}(b), but . \ac{mpjpe} of thumb and index fingertips are worse on \textit{hand-hand} protocol as shown in \figref{pck_curves}(d). We hypothesize this is a disadvantage of directly regressing joint angles that small errors of individual joints can propagate and be amplified along the kinematic chain.

\paragraph{Unknown hand skeleton} When the user's hand skeleton is not provided, we perform a hand scale calibration for both our method and \cite{MeGATrack_Siggraph2020}. Hand scale calibration in \cite{MeGATrack_Siggraph2020} is achieved by gathering multiple multi-view frames and using an optimization method to solve for the hand scale. In our work, we  calibrate the hand scale by averaging scale predictions of the first 30 frames. 
To gather metrics, we first run both methods to perform scale calibration to obtain calibrated hand skeletons. We then re-run inference using the calibrated hand skeletons. In Table \ref{table:compare_previous}, our method consistently outperforms \cite{MeGATrack_Siggraph2020} in \ac{mpjpe} metric by a large margin (11.2mm vs. 12.9mm on \textit{separate-hand}; 12.0mm vs. 13.6mm on \textit{hand-hand}), indicating the effectiveness of our end-to-end differentiable architecture.
% . We hypothesize our method benefits from the end-to-end differentiability where the pose loss directly mirrors the \ac{mpjpe} metric so that our architecture can more easily learn hand size predictions contributes to lower \ac{mpjpe}. In contrast, the keypoint loss used by \cite{MeGATrack_Siggraph2020} has a indirect mapping to the \ac{mpjpe} metric and makes less accurate hand size predictions. 
%We also note that our work is the first to demonstrate effective scale prediction using a neural network to our knowledge.

\subsection{Pinch evaluation}\label{sec:pinch_eva}
\begin{table}[t]
\caption{Comparison on pinch metrics. For each of our models, loss functions used for training are specified after "\textbf{|}".}
\begin{center}
\label{table:pinch}
\renewcommand{\arraystretch}{0.95}
\begin{tabular}{l|ccc}
\hline
model & precision (\%) & recall (\%)\\
\hline
\cite{MeGATrack_Siggraph2020} & 96.5 & 95.0 \\
\hdashline
Ours \textbf{|} $L_{pose}$ + $L_{temp}$                                  & 92.2 & 90.8 \\
Ours \textbf{|} $L_{pose}$ + $L_{temp}$ + $L_{pinch}$                    & \textbf{97.3} & \textbf{97.3} \\
\hline
\end{tabular}
\end{center}
\end{table}
In this section, we compare the precision and recall metrics~\cite{lecun2015deep} for pinch detection.
A pinch detector based on index-thumb fingertip distance thresholding is used for both methods. 
The threshold is the same as used in our designed $L_{pinch}$. 
As shown in Table \ref{table:pinch}, our model trained without $L_{pinch}$ gives worse metrics than \citet{MeGATrack_Siggraph2020}.
This could be related to the observation made in \secref{evaluation:compare_to_previous} that our method is less capable of precise localization.
% To further understand this problem, we plotted the pck curves for the thumb and index fingers in \figref{pck_curves} (b) and (d) and observed the same phenomenon.
% We think our model can potentially be improved by borrowing ideas from multi-stage methods which we'll discuss in \secref{conclusion}.
% When designing $L_{pinch}$, we intentionally use the same distance thresholds as the pinch detector.
%By doing this, our network is directly optimizing for the pinch detection task whereas such task level optimization can't be achieved by a multi-stage method.
After incorporating $L_{pinch}$ for training, though the \ac{mpjpe} is slightly degraded as shown in Table \ref{table:compare_previous}, the pinch detection accuracy is significantly boosted. 
In Table \ref{table:pinch}, our model trained with $L_{pinch}$ outperforms \citet{MeGATrack_Siggraph2020} (97.3\% vs. 96.5\% in precision; 97.3\% vs. 95.0\% in recall). We'd like to emphasize our model provides a framework for studying the interplay between pose estimation and many other downstream tasks such as pinch detection. 
These tasks are much harder to be jointly optimized in multi-stage methods.
\section{Conclusion}
\label{sec:conclusion}
We have presented an end-to-end differentiable architecture designed for hand tracking in VR. 
\revised{It unifies multi-view, temporal fusion and handling of wide \ac{fov} images while making absolute 3D hand pose predictions.} We demonstrate end-to-end differentiability makes it easier to optimize for downstream tasks like jitter reduction and pinch detection compared to multi-stage non-differentiable pipelines. A new large-scale egocentric dataset is introduced. 
We demonstrate compelling VR experiences in the supplementary video with our model trained on this new datset.

\paragraph{Limitations and future work} Hand-hand interactions remain a challenge for our method. A potential solution is joint regression of both hand poses and designing hand-hand interaction losses (i.e. loss that prevents inter-penetration). Compared to \cite{MeGATrack_Siggraph2020}, we discovered our method is less capable of precise localizations. We hypotheize this to be a limitation with the direct pose regression approach. By borrowing ideas from the multi-stage method, our model can potentially incorporate heatmap regression and numerical optimization as differentiable components during training to achieve more precise hand tracking.

\bibliographystyle{ACM-Reference-Format}
\bibliography{references}

\appendix
\section{Network architecture details}
\label{sec:arch}

\begin{table}[]
\caption{Architecture table}
\label{table:arch}
\centering
\renewcommand{\arraystretch}{1} 
\resizebox{\linewidth}{!}{%
\begin{tabular}{@{}ccccc@{}}
\toprule
Module                     & Input                                    & Output                      & Hidden state & Layers \\ \midrule
Encoder                    & $1\times96\times96$            & $72\times6\times6$   & NA & resnet + Conv11       \\
Multi-view fusion      & $144\times6\times6$             & $72\times6\times6$           & NA   & (Conv11 + ReLU) $\times 2$ + Conv11     \\
Temporal module             & $72\times6\times6$              & $72\times6\times6$  & $18\times6\times6$   & (Conv11 + ReLU) $\times 2$ + Conv11  \\
Skeleton encoder      & 120                                       & $4\times6\times6$    & NA   & linear + reshape  \\
Regressor-K           & $76\times6\times6$               & $41$     & NA   & residual blocks $\times$ 2 + Pool \\
Regressor-U            & $72\times6\times6$               & $42$     & NA   & residual blocks $\times$ 2 + Pool \\ \bottomrule
\end{tabular}
}
\end{table}

The input shape, output shape, hidden state shape and the layers used for each module are shown in Table \ref{table:arch}.
The encoder uses the same resnet as \cite{MeGATrack_Siggraph2020} to ensure fair comparisons. The last layer of the encoder is a $1\times1$ convolution layer for dimensionality reduction purpose.
Multi-view fusion uses multiple $1\times1$ convolutions and ReLU layers. Each $1\times1$ convolution serves the purpose of feature fusion and dimensionality reduction. The output shape of the multi-view fusion module is the same as the output shape of the encoder. The temporal module is a recurrent neural network with a hidden state using $1\times1$ convolution and ReLU as the building blocks.
Both Regressor-K and Regressor-U are built from residual blocks. The output of Regressor-K contains 20 dimensional joint angles and 21 dimensional root point coordinates.
Regressor-U outputs a 1 dimensional hand scale parameter in addition to joint angle and root point outputs. 

For root transform prediction, we pre-define $7$ points for representing a transformation in the hand local space: $v_H = \{[0, 0, 0]^T, $ $[1, 0, 0]^T, [0, 1, 0]^T, [0, 0, 1]^T, [1, 1, 0]^T, [1, 0, 1]^T, [0, 1, 1]^T\}$.
And the task of a regressor is to predict the location of these points denoted as $\hat{v}$ in the reference camera space. The root transformation can be recovered using Singular Value Decomposition~\cite{SorkineRabinovich:SVD-rotations:2016} by solving the following equation:
\begin{align}
\hat{T}_H = \min_{\hat{T}_H} \sum_i || \hat{T}_H * v_{H, i} - \hat{v}_i||_2^2
\end{align}

\end{document}